\documentclass{scrartcl}
\usepackage[top=2.5cm, bottom=3cm,footskip=8mm, left=2cm, right=2cm]{geometry}
\usepackage{subcaption}
\usepackage{siunitx}
\usepackage{amsmath}
\usepackage{authblk}
\usepackage{booktabs}
\usepackage[hidelinks,colorlinks=false]{hyperref}
\usepackage{graphicx} 
\usepackage{xcolor}
\xdefinecolor{darkblue}{rgb}{0.0 0.0 0.4}
\graphicspath{{./Graphics}}
\usepackage[abbreviations,symbols,nogroupskip,nonumberlist]{glossaries-extra}

\makeglossaries[abbreviations]
\newabbreviation{ai}{AI}{artificial intelligence}
\newabbreviation{aru}{ARU}{autonomous recording unit}
\newabbreviation{dnn}{DNN}{deep neural network}
\newabbreviation{fft}{FFT}{fast fourier transform}
\newabbreviation{flops}{FLOPs}{floating point operations}
\newabbreviation{mcu}{MCU}{microcontroller unit}
\newabbreviation{pam}{PAM}{passive acoustic monitoring}
\newabbreviation{ram}{RAM}{random-access memory}
\newabbreviation{rom}{ROM}{read-only memory}
\newabbreviation{sgd}{SGD}{stochastic gradient descent}
\DeclareSIUnit{\watthour}{Wh}
\providecommand{\mytitle}{Investigating Target Class Influence on Neural Network Compressibility for Energy-Autonomous Avian Monitoring}
\hypersetup{%
    colorlinks=false, linktocpage=false, pdfborder={0 0 0}, pdfstartpage=3, pdfstartview=FitV,%
    hypertexnames=true, pdfhighlight=/O,
    pdftitle={N.~Brolich et al.: \mytitle},%
    pdfauthor={\textcopyright\ Nina Brolich},%
    pdfsubject={},%
    pdfkeywords={},%
    pdfcreator={pdfLaTeX},%
    pdfproducer={LaTeX}%
}
\title{\mytitle}
\author[1,2,3]{Nina Brolich}
\author[1]{Simon Geis}
\author[1]{Maximilian Kasper}
\author[4]{Alexander Barnhill}
\author[1]{Axel Plinge}
\author[1,5]{Dominik Seuß}

\affil[1]{Fraunhofer Institute for Integrated Circuits IIS, Germany}
\affil[2]{Fachhochschule Erfurt, Germany}
\affil[3]{Universität Erfurt, Germany}
\affil[4]{Friedrich-Alexander-Universität Erlangen-Nürnberg (FAU), Germany}
\affil[5]{Center for Artificial Intelligence and Robotics (CAIRO),
Technische Hochschule Würzburg-Schweinfurt, Germany}

\date{}

\begin{document}
\maketitle
\begin{abstract}
    Biodiversity loss poses a significant threat to humanity, making wildlife monitoring essential for assessing ecosystem health. Avian species are ideal subjects for this due to their popularity and the ease of identifying them through their distinctive songs. Traditional avian monitoring methods require manual counting and are therefore costly and inefficient. In passive acoustic monitoring, soundscapes are recorded over long periods of time. The recordings are analyzed to identify bird species afterwards. Machine learning methods have greatly expedited this process in a wide range of species and environments, however, existing solutions require complex models and substantial computational resources. Instead, we propose running machine learning models on inexpensive \glspl{mcu} directly in the field. Due to the resulting hardware and energy constraints, efficient \gls{ai} architecture is required. 
    
    In this paper, we present our method for avian monitoring on \glspl{mcu}. We trained and compressed models for various numbers of target classes to assess the detection of multiple bird species on edge devices and evaluate the influence of the number of species on the compressibility of neural networks. Our results demonstrate significant compression rates with minimal performance loss. We also provide benchmarking results for different hardware platforms and evaluate the feasibility of deploying energy-autonomous devices.

\end{abstract}

\glsresetall

\section{Introduction}
Among today's global environmental crises, the ongoing loss of biodiversity stands out as especially severe: it ensures access to vital resources \cite{singh_principal_2021} and contributes to various environmental services such as climate regulation, control of pollution and soil erosion, as well as pollination of crops, while also holding intrinsic value for cultural identity \cite{habibullah_impact_2022}. Currently, biodiversity is declining at unprecedented rates, highlighting the need for conservation efforts \cite{almond_living_2020}. 
In this context, biodiversity monitoring provides essential information for tracking environmental changes and plays an integral role in assessing population trajectories of different species \cite{schmeller_bird-monitoring_2012}. Birds are particularly well suited for monitoring due to their frequent and distinctive vocalizations \cite{shonfield_autonomous_2017}, their role as indicators of ecosystem health \cite{mekonen_birds_2017}, and their popular appeal \cite{sekercioglu_promoting_2012}.

Avian monitoring has been done mainly through point counts \cite{cavarzere_recommendations_2013}, where the goal is to record all birds seen or heard within a given time period, with the observer being stationary \cite{etterson_estimating_2009}. However, point counts are highly susceptible to external circumstances, such as weather conditions, dependent on human expertise \cite{brewster_testing_2009}, and their scalability is limited by logistic and financial constraints \cite{kahl_birdnet_2021}.

In recent years, \glspl{aru} have emerged as a cost-effective alternative to point counts, allowing long-term sound collection through \gls{pam} with minimal disturbance, broader spatial and temporal coverage, and permanent recordings for re-analysis \cite{perez-granados_estimating_2021}.
However, analyzing these large datasets is challenging \cite{molina-mora_utility_2024}: manual methods require reducing the number of recordings or species (\cite{furnas_rapid_2020}), while automatic approaches remain difficult. \Glspl{dnn} show promise but face computational constraints (\cite{potamitis_automatic_2014, znidersic_using_2020}). \emph{BirdNET} \cite{kahl_birdnet_2021}, capable of identifying thousands of species, is an example of a successful \gls{dnn}-based solution and provides our method with a baseline for data collection and pre-processing. However, this approach is computationally expensive, requires internet access, and often exceeds the needs of localized surveys.

As an alternative, we propose species identification in real-time on an embedded, energy self-sustaining solution, shifting from retrospective analysis of \gls{aru} data to in-field processing, thereby reducing computational overhead while improving both energy and cost efficiency. The system is to be realized with edge \gls{ai} using a \gls{mcu}, which introduces resource constraints such as limited memory, low processing capacity, and lack of parallelism \cite{sudharsan_ml-mcu_2021}. To operate effectively under these constraints, deployment on edge devices like \glspl{mcu} requires compact models, achieved either through lightweight network design or compression, to accommodate hardware constraints while preserving performance.

To assess the feasibility of avian monitoring on the edge and to examine target class influence on compressibility, we trained and compressed models for various numbers of avian target classes and benchmarked their energy consumption and latency on an \emph{ARM Cortex-M4}, an \emph{ARM Cortex-M7}, and a \emph{Raspberry Pi 4}. Based on the results, we provide an estimate for the required battery capacity for real-life deployment.
In the following, we describe our methodology, present and discuss the results, and conclude with an outlook on future work.

\section{Methodology}
\begin{figure}
    \centering
    \includegraphics[width=0.8\textwidth]{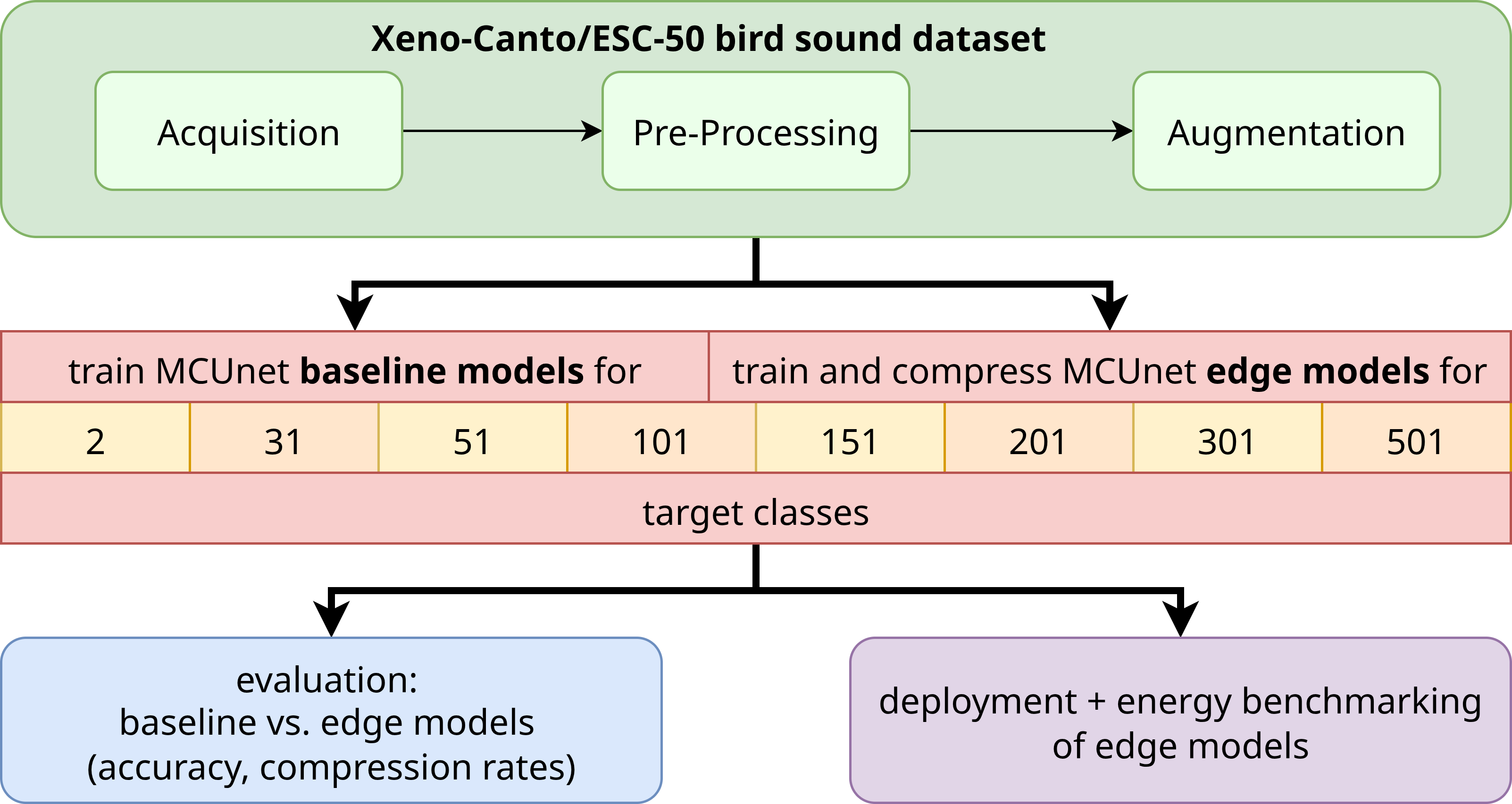}
    \caption{Methodology overview: workflow for dataset preparation, MCUNet model
training, compression, and edge benchmarking.}
    \label{fig:overview}
\end{figure}

An overview of our methodology is provided in Fig.~\ref{fig:overview}.
The dataset was primarily compiled from \emph{Xeno-Canto}~\cite{planque_xeno-canto_2005}, an open database of user-submitted bird recordings, resulting in a large but heterogeneous collection in both quality and duration. 500 species were selected based on the availability of sufficient samples, with preference first given to German, then European and finally global species. German species and data were
prioritized to match the intended deployment region and to ensure data
representativeness. For each species, 250 recordings were randomly chosen. In addition, the \emph{ESC-50}~\cite{piczak_esc_2015} dataset was used to form a single non-avian class by merging 49 environmental sound categories (excluding bird calls).
The data samples were pre-processed using the following steps:
\begin{enumerate}
    \item Discarding audio samples shorter than 2 seconds, based on the average length of a bird vocalization of 1.94 seconds \cite{kahl_identifying_2019}.
    \item Removing silent sections, defined as amplitudes below \qty{20}{\percent} of the maximum peak.
    \item Sequentially splitting recordings into 2-second chunks with a maximum of 30 chunks per recording while discarding chunks without peaks (at least \qty{7.5}{\percent} louder than surrounding points).
    \item Normalizing each chunk so that its maximum absolute amplitude equals 1 to ensure consistency in amplitude throughout the dataset.
    \item Converting the normalized chunks into mel-spectrograms, using a sample rate of \qty{48}{\kilo\hertz}, 64 mel bands, a \gls{fft} window size of 512 and a hop length of 384~\cite{kahl_birdnet_2021}. The frequency was constrained to \qty{150}{\hertz} and \qty{7.5}{\kilo\hertz}.
\end{enumerate}

This produced an easily attainable but heterogeneous dataset, containing
samples of varying quality with both avian and non-avian background
noises. While not optimal for training, it realistically represents the acoustic
challenges encountered in edge-deployed avian monitoring. On average, this
resulted in 2452 chunks per bird species with a standard deviation of 874.
The dataset was partitioned into training, validation, and test sets without
data leakage.

\begin{figure}
    \centering
    \begin{subfigure}[t]{0.45\textwidth}
        \centering
        \includegraphics[width=\textwidth]{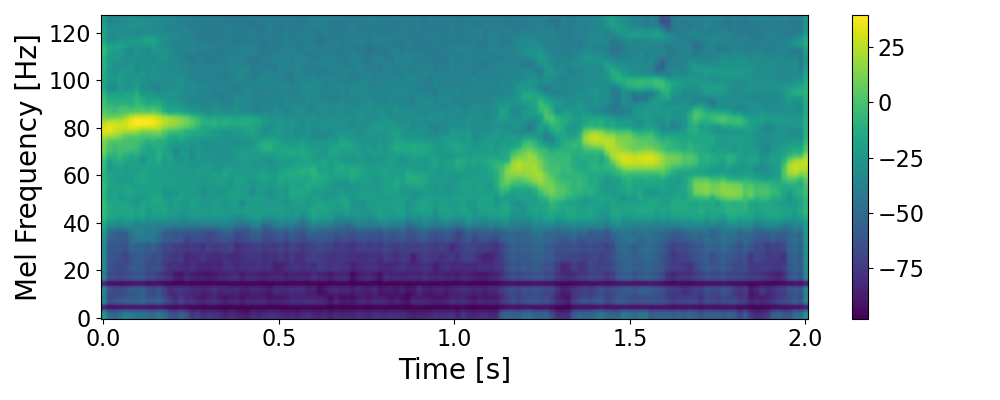}
        \captionsetup{width=.8\linewidth}
        \caption{Original input mel-spectrogram}
    \end{subfigure}%
    \begin{subfigure}[t]{0.45\textwidth}
        \centering
        \includegraphics[width=\textwidth]{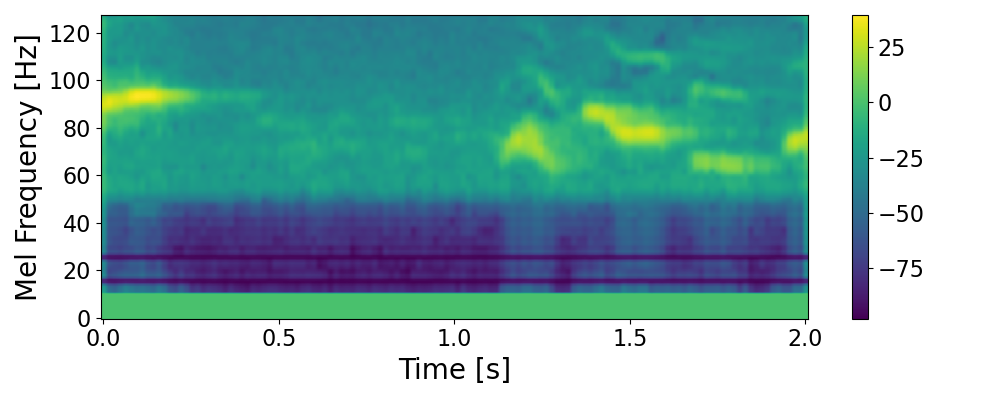}
        \captionsetup{width=.8\linewidth}
        \caption{Vertically shifted mel-spectrogram}
    \end{subfigure}
    
    \begin{subfigure}[t]{0.45\textwidth}
        \centering
        \includegraphics[width=\textwidth]{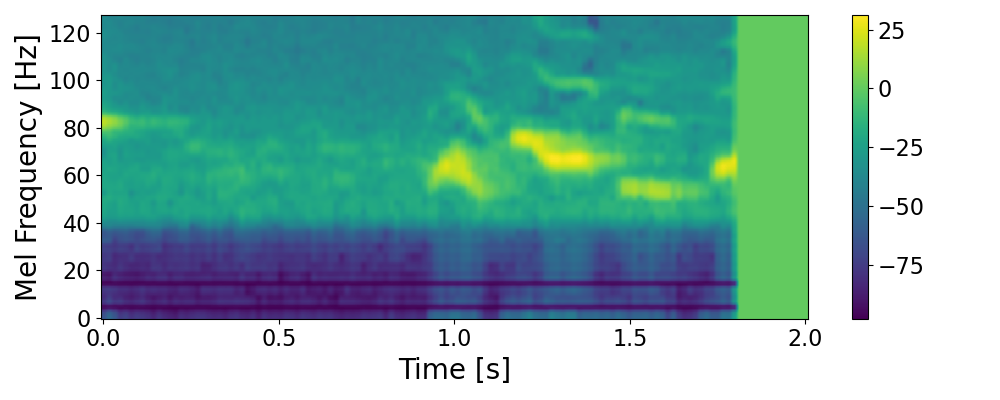}
        \captionsetup{width=.8\linewidth}
        \caption{Horizontally shifted mel-spectrogram}
    \end{subfigure}%
    \begin{subfigure}[t]{0.45\textwidth}
        \centering
        \includegraphics[width=\textwidth]{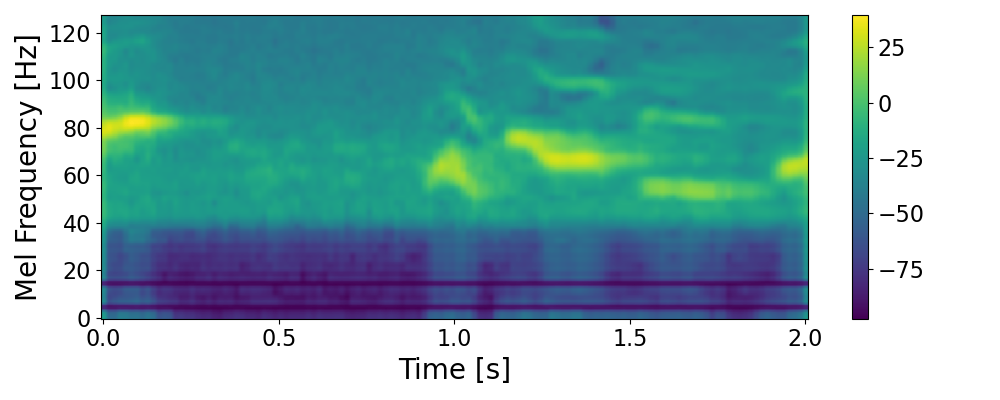}
        \captionsetup{width=.8\linewidth}
        \caption{Time-warped mel-spectrogram}
    \end{subfigure}
    
    \begin{subfigure}[t]{0.45\textwidth}
        \centering
        \includegraphics[width=\textwidth]{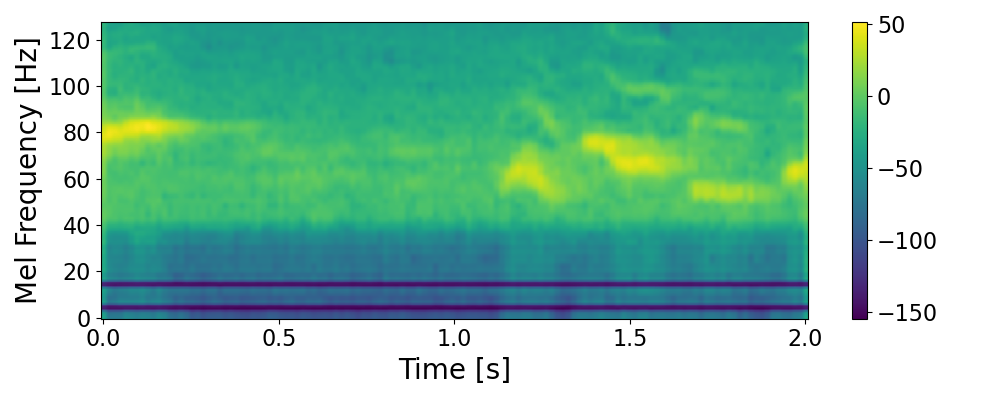}
        \captionsetup{width=.8\linewidth}
        \caption{Mel-spectrogram with added noise}
    \end{subfigure}%
    \begin{subfigure}[t]{0.45\textwidth}
        \centering
        \includegraphics[width=\textwidth]{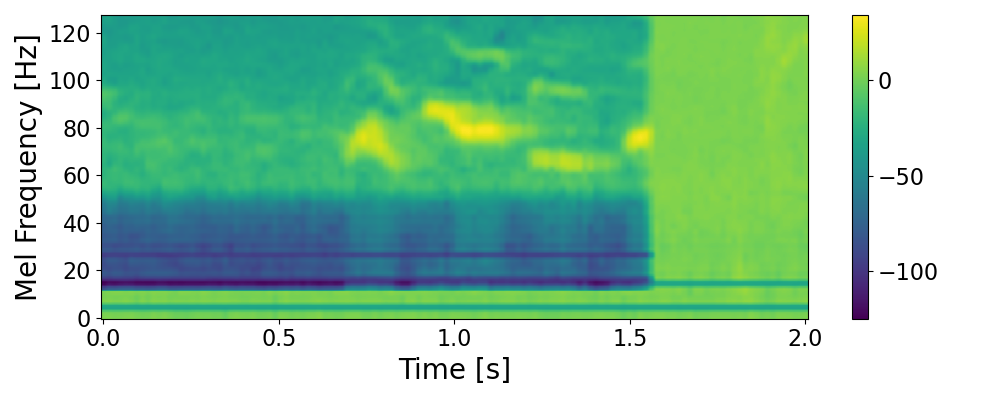}
        \captionsetup{width=.8\linewidth}
        \caption{Mel-spectrogram with all augmentations}
    \end{subfigure}
    \caption[Data augmentation mel-spectrograms]{Mel-spectrograms with different data augmentation methods applied.}
    \label{fig:data-augmentations}
\end{figure}

To increase the diversity of our training data and to enhance the generalizability of our models, we applied the data augmentation pipeline outlined by \cite{kahl_birdnet_2021} to our data.
Four different augmentation methods were applied, as illustrated in Fig.~\ref{fig:data-augmentations}.

\begin{itemize}
    \item \textbf{Random vertical shifts (frequency roll):} the mel-spectrogram was shifted up or down by a random factor between -\qty{5}{\percent} and \qty{5}{\percent}, with the vacated area filled with zeros.
    \item \textbf{Random horizontal shifts (time roll):} the mel-spectrogram was shifted left or right by a random factor between -\qty{25}{\percent} and \qty{25}{\percent}, with zeros filling the empty region.
    \item \textbf{Time warping:} the mel-spectrogram was deformed in time direction using the \emph{SpecAugment} algorithm \cite{park_specaugment_2019}.
    \item \textbf{Addition of noise:} randomly selected noise chunks, previously discarded during preprocessing and verified to contain no bird calls, were added at a random intensity between \qty{20}{\percent} and \qty{80}{\percent}.
\end{itemize}

 These augmentation methods all represent acoustic variations between training and test data like the changes in the vocal output of birds depending on environmental factors, high levels of ambient noise, and lack of training sample diversity \cite{kahl_birdnet_2021}. Each augmentation was applied with a \qty{50}{\percent} probability per chunk, in random order, with a maximum of three augmentations per chunk.\\

\begin{figure}
    \centering
    \includegraphics[width=\textwidth]{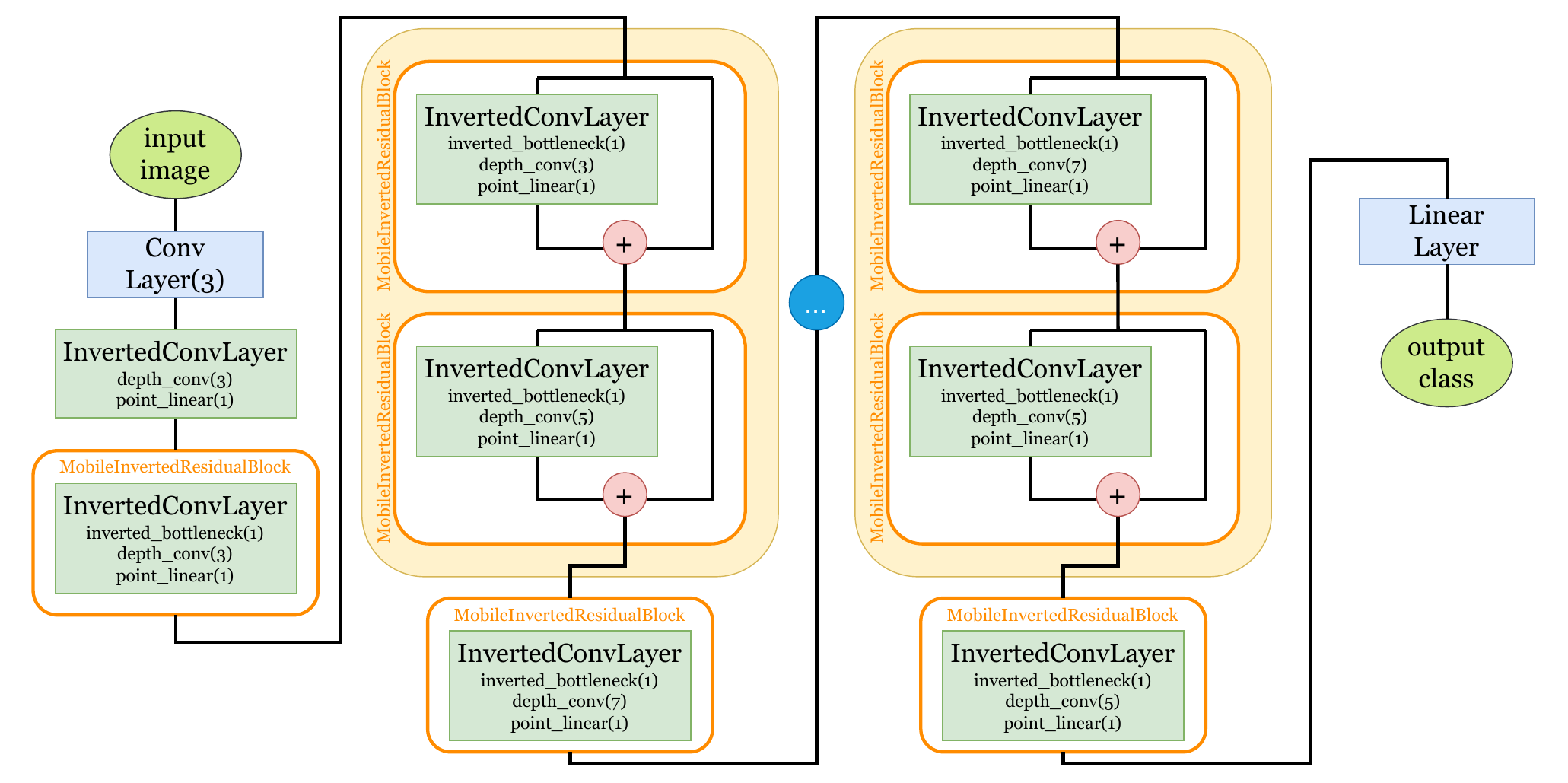}
    \caption{Structure and components of \emph{MCUNet}.}
    \label{fig:mcunet-architecture}
\end{figure}

For the experiments, the \emph{mcunet-in4} model from the \emph{MCUNet} framework \cite{lin_mcunet_2020} was selected and adapted for bird sound classification. \emph{MCUNet} is specifically designed for deep learning on microcontrollers, combining efficient neural architecture search (\emph{TinyNAS}) with a memory-optimized inference engine (\emph{TinyEngine}) to accommodate hardware constraints. Despite the existance of smaller \emph{MCUNet} models, \emph{mcunet-in4} was chosen for its better performance and to ensure support for higher numbers of target classes. Nevertheless, it still allows deployment on edge devices, which was also aided by further compression of the models. Its architecture, illustrated in Fig.~\ref{fig:mcunet-architecture}, consists of an initial convolutional layer, 17 \emph{MobileInvertedResidualBlocks} \cite{sandler_mobilenetv2_2018}, and a final linear layer. To accommodate mel-spectrogram inputs, the first layer was modified to accept a single input channel, while the output layer was adjusted to match the number of target classes (31, 51, 101, etc.), including a non-avian class. Pre-trained weights from \emph{ImageNet}~\cite{deng_imagenet_2009} were loaded for all layers except the first and last.

In total, one uncompressed baseline model and 50 compressed edge models were trained for 2, 31, 51, 101, 151, 201, 301, and 501 target classes. For the model with only two target classes, the objective was to identify one bird species against 29 others, as well as non-bird data, i.e., it was trained with two classes, one consisting of data for the common blackbird (\emph{turdus merula}), and the other one consisting of data of the 29 other most common bird species in Germany, as well as the environment data. From 31 target classes onward, the models were trained to identify 30 (or 50, 100, \dots) bird species, as well as one non-event class.

The baseline models were trained with a set number of 30 training epochs. The training was conducted with a batch size of 32, using the \gls{sgd} optimizer with a learning rate of 0.001 and a momentum of 0.9. Cross-entropy loss served as the loss function. The edge models were trained using an internal tool from \emph{Fraunhofer IIS}. While the same general training configuration was utilized as for the baseline models, an interleaved pruning process with quantization at the end of the training process was employed. This tool implements the neural network compression approach from \cite{deutel_combining_2024}, performing multiple training and compression trials to maximize accuracy and minimize \gls{rom}, \gls{ram}, and \gls{flops} requirements. For each edge model, 50 trials were run, producing 50 compressed models, some of which are Pareto optimal across these objectives. A model is considered Pareto optimal if no other model performs better in one objective without performing worse in at least one other.
\\

For the evaluation of the results, the performance of the edge models was first compared to the corresponding baseline models. A representative compressed model was selected from the 50 trials conducted for each target class number. To rank the trial $t_x$, the trade-off between accuracy $acc(t_x)$ (Eq.~\ref{ranking-trials:accuracy}) and the combined memory metrics $mem(t_x)$ of \gls{rom}, \gls{ram}, and \gls{flops} (Eq.~\ref{ranking-trials:memory}) was emphasized, as showcased in the ranking function $r(t_x)$ in Eq.~\ref{ranking-trials:rank}.
\begin{equation}
    \label{ranking-trials:accuracy}
    acc(t_x) =\frac{\text{ACC}(t_x)}{\max\{\text{ACC}(t_i) | i \in [0, 49]\}}
\end{equation}

\begin{equation}
    \label{ranking-trials:memory}
    mem(t_x) = \frac{
    1 - \frac{\text{RAM}(t_x)}{\max\{\text{RAM}(t_i)\}} + 
    1 - \frac{\text{ROM}(t_x)}{\max\{\text{ROM}(t_i)\}} + 
    1 - \frac{\text{FLOPs}(t_x)}{\max\{\text{FLOPs}(t_i)\}}
    }
    {3},\ i \in [0, 49]
\end{equation}
\begin{equation}
    \label{ranking-trials:rank}
    r(t_x) = mem(t_x) + acc(t_x)
\end{equation} 

To evaluate compressibility, we defined \gls{rom}, \gls{ram}, and \gls{flops} compression rates $cr$, shown in Eq.~\ref{eqn:3} for \gls{flops}, with the values for \gls{rom} and \gls{ram} computed analogously.
\begin{equation}
    \label{eqn:3}
     cr_{\text{FLOPs}}(\text{FLOPs}_{\text{baseline}}, \text{FLOPs}_{\text{edge}}) = 1 - \frac{\text{FLOPs}_{\text{edge}}}{\text{FLOPs}_{\text{baseline}}}
\end{equation}

The overall compression rate of a model was defined as the mean of its \gls{flops}, \gls{rom}, and \gls{ram} compression rates, while the average overall compression rate was computed as the mean of these values across all Pareto-optimal trials for the target classes.\\

The \emph{dnnruntime} framework \cite{deutel_energy-efficient_2023} was used to convert the models to C code and create a deployable binary file. 

This binary file was then flashed onto \emph{ARM Cortex-M4} and \emph{ARM Cortex-M7} processors on a \emph{SparkFun MicroMod ATP} carrier board to measure the latency and energy consumption of the models on the two microcontrollers.
Additionally, we benchmarked the models on a \emph{Raspberry Pi 4} using the \emph{onnxruntime} framework~\cite{onnxruntime}. As the inference on this device may be influenced by the operating system and the scheduler, we measured 1000 inference steps and provide the mean and standard deviation of the results.

For an estimation of the feasibility of energy-autonomous deployment, we make the following assumptions:
\begin{enumerate}
    \item The device is equipped with a battery which should be able to power the device for at least 48 hours without charging.
    \item In addition, the device is equipped with a solar panel that should be able to fully charge the battery in 24 hours.
    \item The device is in sleep mode per default and wakes up every 10 seconds. If the device recognizes sound from the microphone, inference is performed as long as sound is detected. Otherwise, the device returns to sleep for the next 10 seconds. We assume that \qty{10}{\percent} of the time, the inference step is performed.
\end{enumerate}

Based on the power required during inference and idle, we can then infer the required battery size of the device for running for 48 hours. The required size $A$ of the solar cell can then be estimated by the average sun-radiation power density $S_\text{rad}$ in Germany and by the required charging output $P_\text{charge}$ as shown in Eq.~\ref{eq:area}. We assume a solar panel efficiency $\eta_\text{solar}$ of \qty{20}{\percent} and a charging efficiency $\eta_\text{bat}$ of \qty{90}{\percent}.
\begin{equation}
    \label{eq:area}
    A =  \dfrac{P_\text{charge}}{\eta_\text{solar} \cdot \eta_\text{bat} \cdot S_\text{rad}}
\end{equation}

\section{Results}

\begin{figure}
\centering
\includegraphics[width=.6\linewidth]{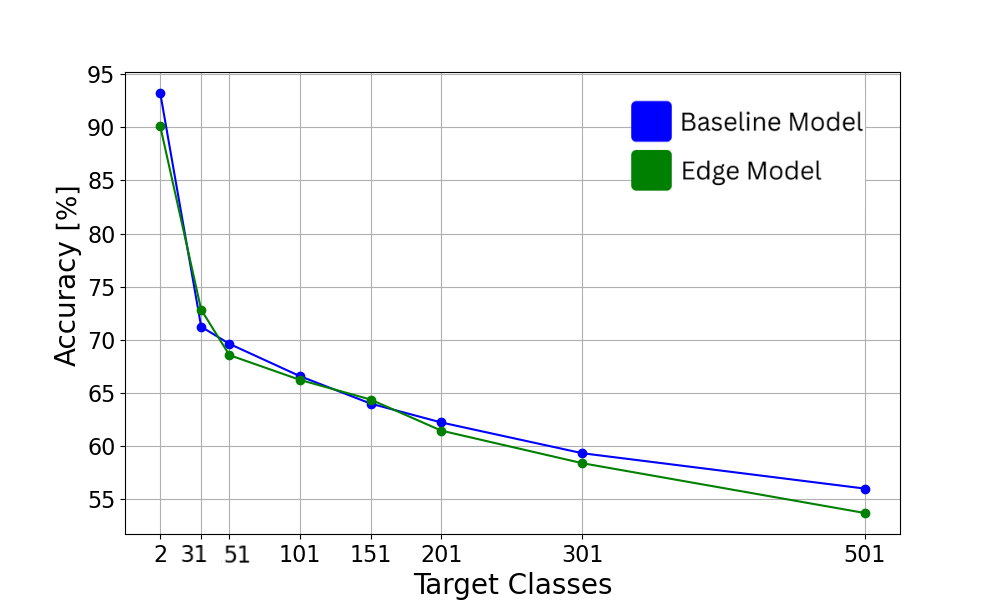}
\caption{Validation accuracies for one baseline and one edge model per number of target classes.}
\label{fig:accuracies}
\end{figure}    
\begin{figure}
\centering
\includegraphics[width=.6\linewidth]{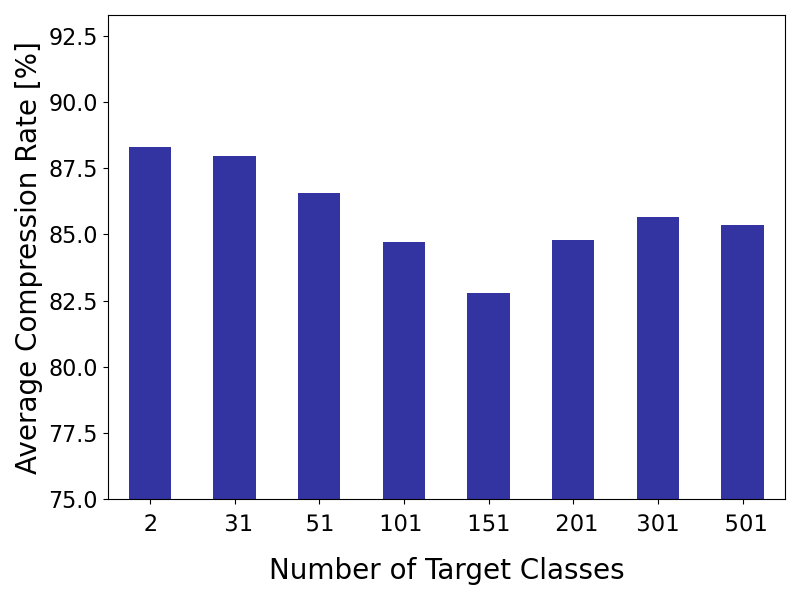}
\caption{Average overall compression rate for all Pareto optimal trials per number of target classes regarding the reduction in \gls{ram}, \gls{rom} and \gls{flops}.}
\label{fig:avg-compression-rate}
\end{figure}    

To assess the general performance of the models, one baseline and one representative edge model per target class number were selected according to the rank defined in Eq.~\ref{ranking-trials:rank}. Their validation accuracies are shown in Fig.~\ref{fig:accuracies}.

Overall, validation accuracy declines as the number of target classes increases. While the models achieve high accuracy values for lower target class numbers, performance is more moderate with more target classes. Notably, there is almost no accuracy loss due to compression, as the accuracies of the baseline and edge models remain within a very similar range.

To assess target class influence on compressibility, the average compression rate was computed across different target class numbers and is shown in Fig.~\ref{fig:avg-compression-rate}. The results indicate a slight decline in compressibility with increasing class numbers. From 201 classes onward, however, this trend reverses, with compressibility improving for larger class counts. This result is unexpected, as a larger number of target classes would hypothetically require more complex architectures with higher memory and \gls{flops} demands, thereby reducing compressibility. While this trend is observed up to a certain point, it appears to reverse for larger class counts. Since the training and compression procedures are essentially a black box, a definitive explanation is difficult. Nonetheless, several factors may contribute: with more classes, models may exploit feature sharing more effectively, capturing overlapping features with fewer parameters \cite{liu_visualizing_2019}. Moreover, the inclusion of additional classes may encourage the learning of more generalized and compact representations, ultimately reducing resource requirements \cite{zhu_learning_2022}. However, the observed decrease and subsequent increase in compressibility are subtle, with compression rates ranging only between approximately \qty{82}{\percent} and \qty{88}{\percent} and thus may not reflect a clear or consistent trend.\\

\begin{figure}
\centering
\begin{subfigure}[t]{0.5\textwidth}
    \centering
    \captionsetup{width=.9\linewidth}
    \includegraphics[width=\textwidth]{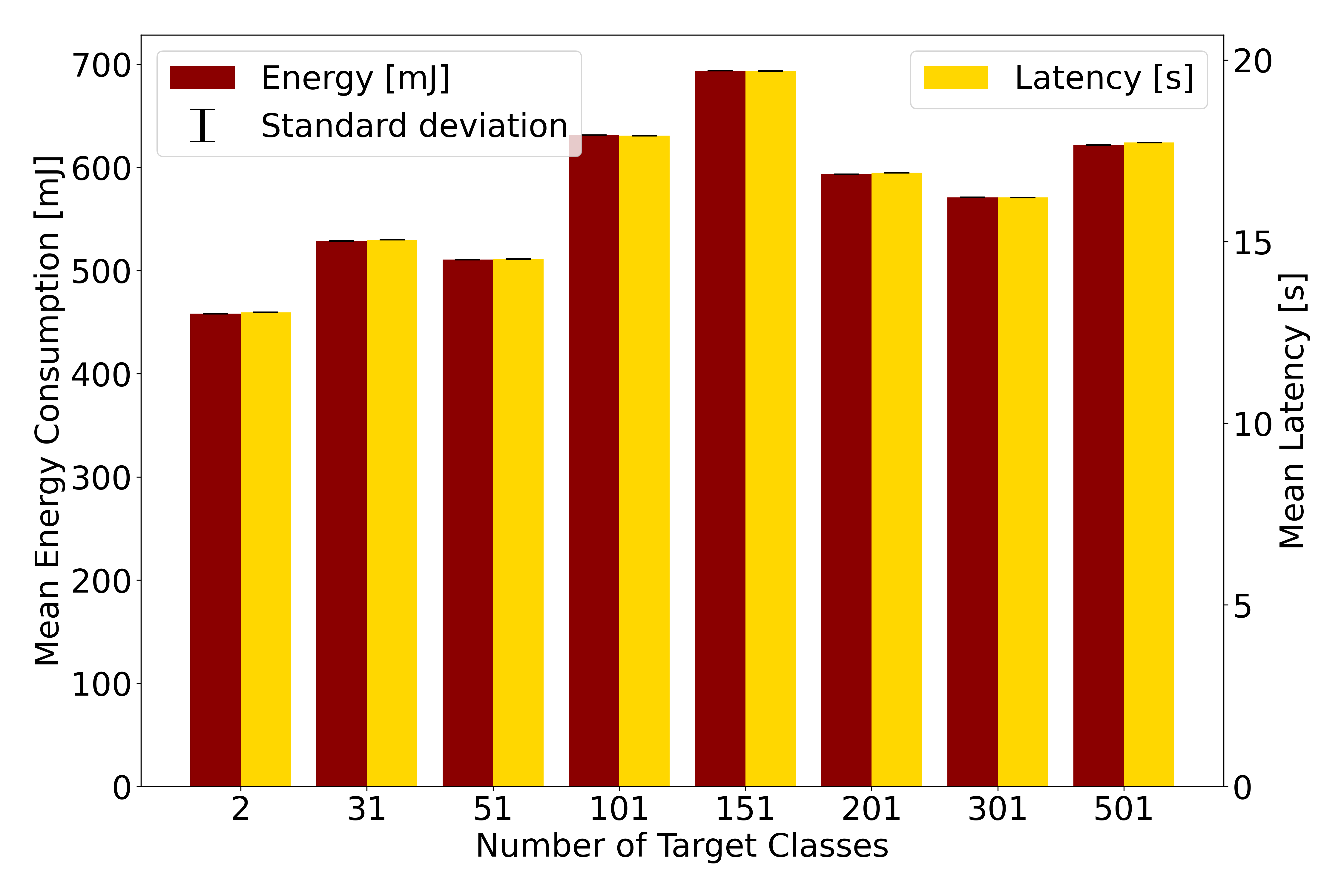}
    \caption{ARM Cortex-M4}
\end{subfigure}%
\begin{subfigure}[t]{0.5\textwidth}
    \centering
    \captionsetup{width=.9\linewidth}
    \includegraphics[width=\textwidth]{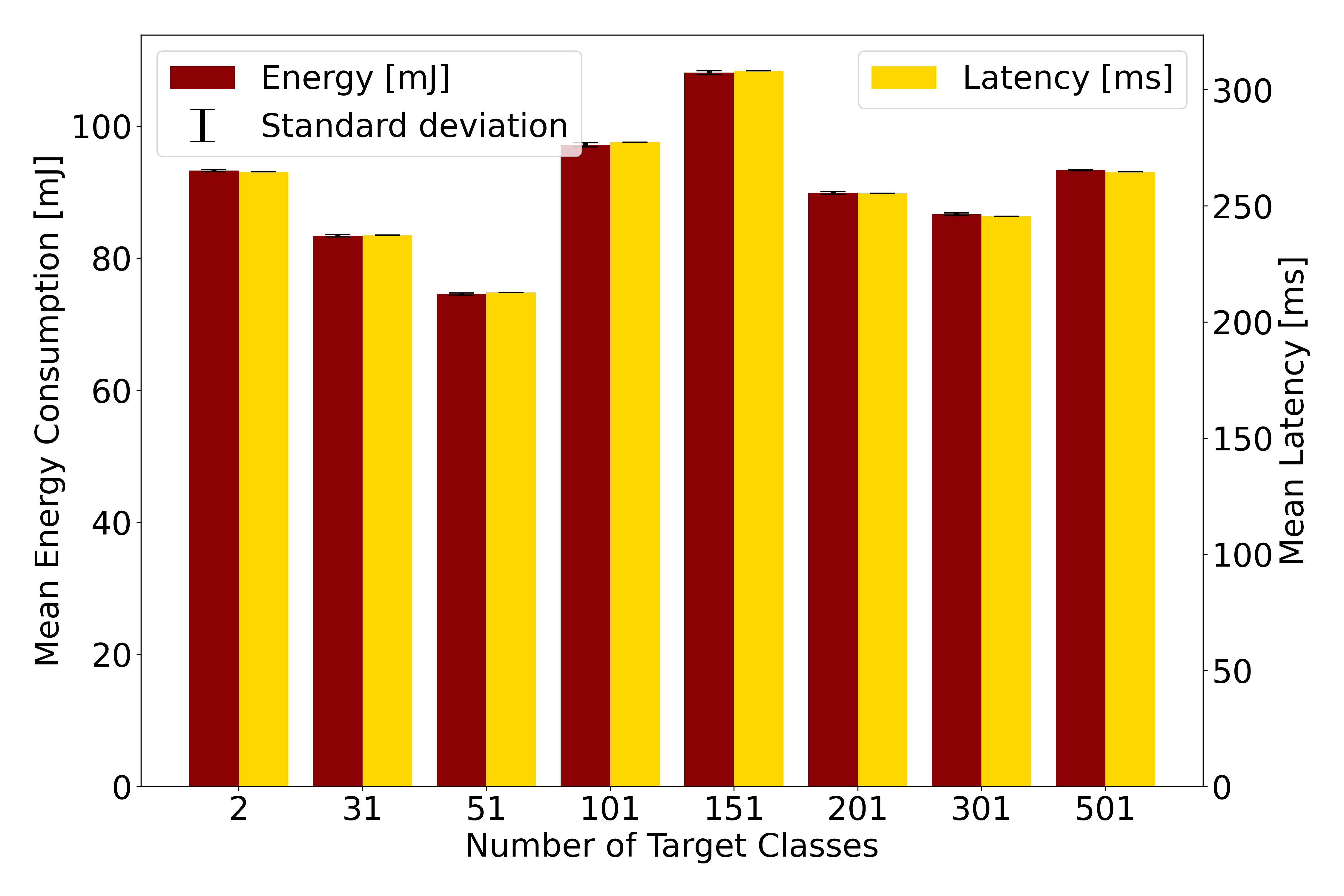}
    \caption{ARM Cortex-M7}
\end{subfigure}

\begin{subfigure}[t]{0.5\textwidth}
    \centering
    \captionsetup{width=.9\linewidth}
    \includegraphics[width=\textwidth]{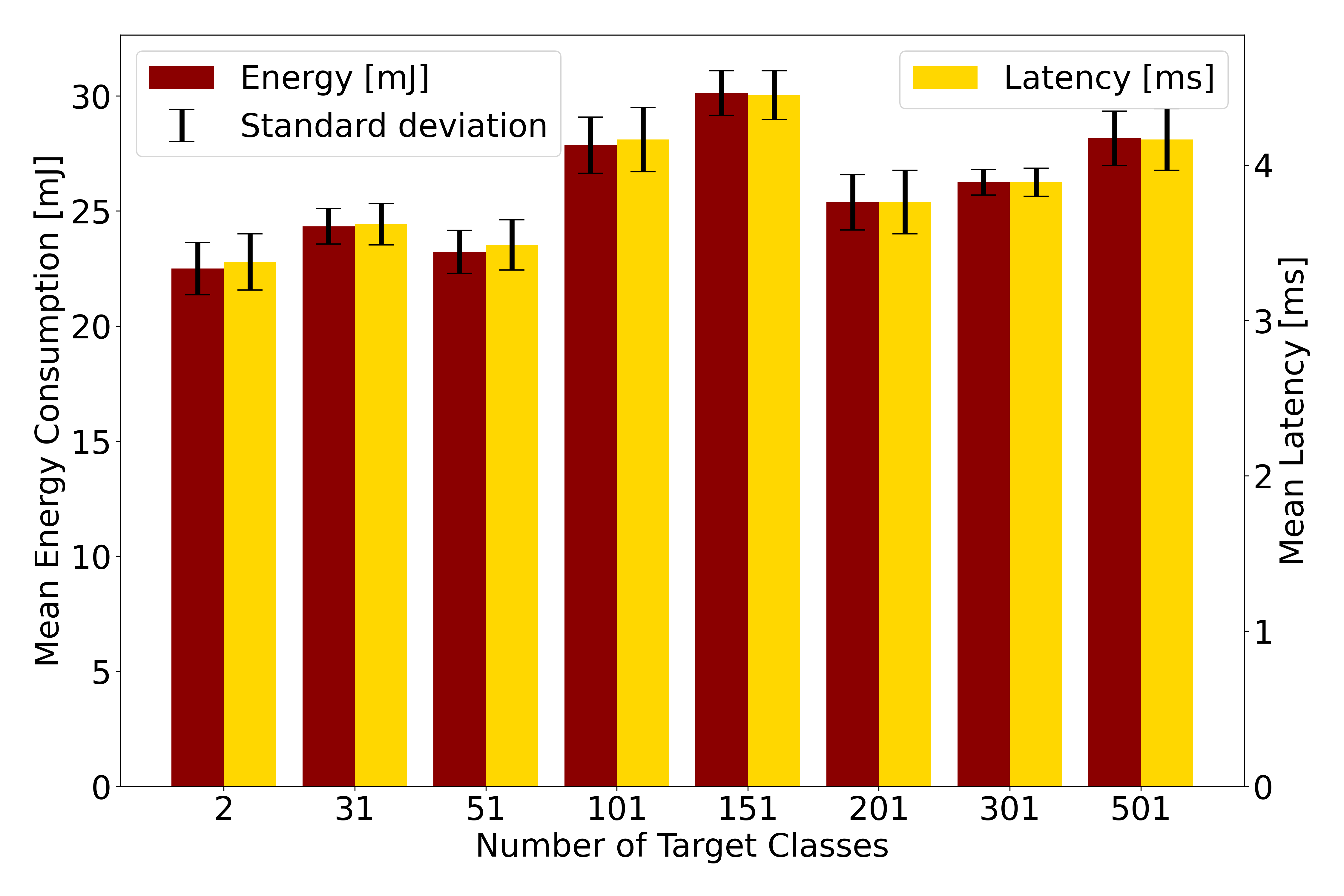}
    \caption{Raspberry Pi 4}
\end{subfigure}
\caption{Average energy consumption and latency of one inference step of the best ranked model for each number of target classes.}
\label{fig:benchmarking}
\end{figure}

To evaluate the feasibility of deploying the compressed models on an \gls{mcu}, energy consumption and latency were measured for one model per number of target classes, selected according to the ranking in Eq.~\ref{ranking-trials:rank}. The results are shown in Fig.~\ref{fig:benchmarking}.
Overall, both latency and energy consumption increase with a growing number of target classes, though the improved compressibility of models with more than 151 classes is partially reflected in lower values for these metrics. Because both energy consumption and latency are largely determined by the number of \gls{flops} performed during inference, they are only indirectly related to compressibility. Since model selection accounted for \gls{flops}, \gls{rom}, and \gls{ram}, cases arise where a model with more \gls{flops} than its successor was chosen due to lower memory requirements. This explains, for example, why the 31-class model consumes more energy than the 51-class model. 
Finally, the benchmarking results indicate that the compressed models achieve energy and latency values suitable for real-world deployment on the \emph{ARM Cortex-M7} and \emph{Raspberry Pi 4}. In contrast, on the more resource-constrained \emph{ARM Cortex-M4}, latency consistently exceeded the audio chunk length, making real-world deployment infeasible.\\

\begin{figure}[ht]
    \centering
    \includegraphics[width=\textwidth]{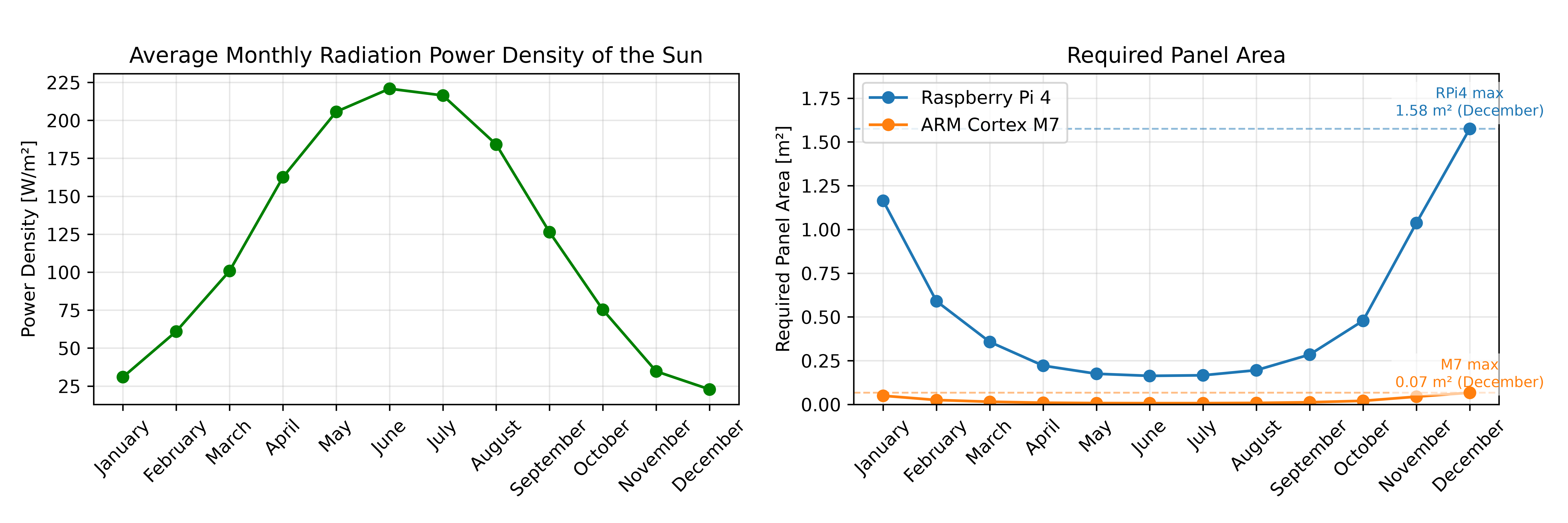}
    \caption{Average power density of the sun's radiation in Germany~\cite{dwd_sun} (left) and the resulting area of the solar panel for each month (right).}
    \label{fig:solar_area}
\end{figure}

For the evaluation of energy-autonomous devices, we selected the model with 31 classes as an example. We measured an energy consumption of \qty{83}{\milli\joule} and a latency of \qty{237}{\milli\second} for the inference and \qty{55}{\milli\joule} and \qty{170}{\milli\second} for the spectrogram generation on the \emph{ARM Cortex M7}. This results in an average power draw of \qty{339}{\milli\watt}. During sleep, we measured a power consumption of \qty{116}{\milli\watt}, resulting in average power consumption of \qty{138.3}{\milli\watt} during the course of the day. Overall, this leads to a required battery capacity of \qty{6.6}{\watthour}. 

For the \emph{Raspberry Pi 4}, we measured \qty{24.3}{\milli\joule} and \qty{3.6}{\milli\second} for the inference and \qty{483.}{\milli\joule} and \qty{80.9}{\milli\second} for the mel-spectrogram generation, leading to an average power of \qty{6.0}{\watt}. During idle, we measured a power consumption of \qty{2.93}{\watt}, resulting in an average power of \qty{3.24}{\watt}. In total, this would result in a required battery capacity of \qty{155.5}{\watthour}.

Based on these values, the required size of the solar panel can be calculated as described in Eq.~\ref{eq:area}. During December, the sun has the lowest power density of \qty{22.8}{\watt\per\metre\squared} as shown in Fig.~\ref{fig:solar_area}. With a required charging power of \qty{275}{\milli\watt}, this results in a required panel area of \qty{0.067}{\meter\squared} for the \emph{ARM Cortex M7}. The \emph{Raspberry Pi} requires a charging power of \qty{6.48}{\watt} and therefore a panel size of \qty{1.58}{\meter\squared}.


\section{Conclusion}
Avian species identification is a challenging task, with model accuracy generally decreasing as the number of target classes increases. This is partly due to the heterogeneity of the dataset and the use of a single, relatively simple model architecture across all class numbers, which reflects realistic conditions for edge deployment and was necessary to facilitate comparability of compressibility across different target class numbers.

Our results demonstrate that high compression rates are achievable with minimal loss of accuracy across different numbers of target classes. Although compressibility initially decreased with increasing class numbers and later increased beyond 151 classes, the overall variation is minor and does not necessarily suggest a clear trend. Instead, it reflects the complex interplay between task complexity, model architecture, and learned representations.
The evaluation of energy consumption and latency shows that real-world deployment is feasible on the \emph{Raspberry Pi 4} and on the \emph{ARM Cortex-M7}, but not on the more constrained \emph{ARM Cortex-M4}, emphasizing the importance of selecting appropriate hardware for edge applications.

Overall, we showed that neural network compression is a practical and effective strategy for edge-based avian monitoring, even for large numbers of target classes. Finally, our results indicate that avian monitoring is feasible on energy-autonomous edge devices which could play a crucial role in wildlife monitoring and biodiversity conservation.

Future work could include curating a scientific dataset, exploring different species arrangements in the training data, comparing alternative compression frameworks, and investigating end-to-end audio models to reduce pre-processing overhead.
Furthermore, real-life deployment will require additional functionalities, such as counting, data storage, and energy management, potentially combining edge \gls{ai} with automated data collection technologies.

\section*{Acknowledgment}
This work is part of the GreenICT@FMD project and is funded by the
German Federal Ministry for Research, Technology and Space (BMFTR)
(grant number 16ME0491K).


\bibliographystyle{IEEEtran}
\newpage
\bibliography{references}

\begin{thebibliography}{10}
\providecommand{\url}[1]{#1}
\csname url@samestyle\endcsname
\providecommand{\newblock}{\relax}
\providecommand{\bibinfo}[2]{#2}
\providecommand{\BIBentrySTDinterwordspacing}{\spaceskip=0pt\relax}
\providecommand{\BIBentryALTinterwordstretchfactor}{4}
\providecommand{\BIBentryALTinterwordspacing}{\spaceskip=\fontdimen2\font plus
\BIBentryALTinterwordstretchfactor\fontdimen3\font minus \fontdimen4\font\relax}
\providecommand{\BIBforeignlanguage}[2]{{%
\expandafter\ifx\csname l@#1\endcsname\relax
\typeout{** WARNING: IEEEtran.bst: No hyphenation pattern has been}%
\typeout{** loaded for the language `#1'. Using the pattern for}%
\typeout{** the default language instead.}%
\else
\language=\csname l@#1\endcsname
\fi
#2}}
\providecommand{\BIBdecl}{\relax}
\BIBdecl

\bibitem{singh_principal_2021}
V.~Singh, S.~Shukla, and A.~Singh, ``The principal factors responsible for biodiversity loss,'' \emph{Open Journal of Plant Science}, vol.~6, no.~1, pp. 011--014, 2021.

\bibitem{habibullah_impact_2022}
M.~S. Habibullah, B.~H. Din, S.-H. Tan, and H.~Zahid, ``Impact of climate change on biodiversity loss: global evidence,'' \emph{Environmental Science and Pollution Research}, vol.~29, no.~1, pp. 1073--1086, 2022, publisher: Springer.

\bibitem{almond_living_2020}
R.~E. Almond, M.~Grooten, and T.~Peterson, \emph{Living {Planet} {Report} 2020-{Bending} the curve of biodiversity loss}.\hskip 1em plus 0.5em minus 0.4em\relax World Wildlife Fund, 2020.

\bibitem{schmeller_bird-monitoring_2012}
D.~Schmeller, K.~Henle, A.~Loyau, A.~Besnard, and P.-Y. Henry, ``Bird-monitoring in {Europe}–a first overview of practices, motivations and aims,'' \emph{Nature Conservation}, vol.~2, pp. 41--57, 2012, publisher: Pensoft Publishers.

\bibitem{shonfield_autonomous_2017}
J.~Shonfield and E.~M. Bayne, ``Autonomous recording units in avian ecological research: current use and future applications.'' \emph{Avian Conservation \& Ecology}, vol.~12, no.~1, 2017.

\bibitem{mekonen_birds_2017}
S.~Mekonen, ``Birds as biodiversity and environmental indicator,'' \emph{Indicator}, vol.~7, no.~21, 2017.

\bibitem{sekercioglu_promoting_2012}
{\c C}.~H. {\c S}ekercioğlu, ``Promoting community-based bird monitoring in the tropics: {Conservation}, research, environmental education, capacity-building, and local incomes,'' \emph{Biological Conservation}, vol. 151, no.~1, pp. 69--73, 2012, publisher: Elsevier.

\bibitem{cavarzere_recommendations_2013}
V.~Cavarzere, G.~P. Moraes, J.~J. Roper, L.~F. Silveira, and R.~J. Donatelli, ``Recommendations for monitoring avian populations with point counts: a case study in southeastern {Brazil},'' \emph{Papéis Avulsos de Zoologia}, vol.~53, pp. 439--449, 2013, publisher: SciELO Brasil.

\bibitem{etterson_estimating_2009}
M.~A. Etterson, G.~J. Niemi, and N.~P. Danz, ``Estimating the effects of detection heterogeneity and overdispersion on trends estimated from avian point counts,'' \emph{Ecological Applications}, vol.~19, no.~8, pp. 2049--2066, 2009, publisher: Wiley Online Library.

\bibitem{brewster_testing_2009}
J.~P. Brewster and T.~R. Simons, ``Testing the importance of auditory detections in avian point counts,'' \emph{Journal of field ornithology}, vol.~80, no.~2, pp. 178--182, 2009, publisher: Wiley Online Library.

\bibitem{kahl_birdnet_2021}
S.~Kahl, C.~M. Wood, M.~Eibl, and H.~Klinck, ``{BirdNET}: {A} deep learning solution for avian diversity monitoring,'' \emph{Ecological Informatics}, vol.~61, p. 101236, 2021, publisher: Elsevier.

\bibitem{perez-granados_estimating_2021}
C.~Pérez-Granados and J.~Traba, ``Estimating bird density using passive acoustic monitoring: a review of methods and suggestions for further research,'' \emph{Ibis}, vol. 163, no.~3, pp. 765--783, 2021, publisher: Wiley Online Library.

\bibitem{molina-mora_utility_2024}
I.~Molina-Mora, V.~Ruíz-Gutierrez, {\' A}.~Vega-Hidalgo, and L.~Sandoval, ``The utility of passive acoustic monitoring for using birds as indicators of sustainable agricultural management practices,'' \emph{Frontiers in Bird Science}, vol.~3, p. 1386759, 2024, publisher: Frontiers Media SA.

\bibitem{furnas_rapid_2020}
B.~J. Furnas, ``Rapid and varied responses of songbirds to climate change in {California} coniferous forests,'' \emph{Biological Conservation}, vol. 241, p. 108347, 2020, publisher: Elsevier.

\bibitem{potamitis_automatic_2014}
I.~Potamitis, S.~Ntalampiras, O.~Jahn, and K.~Riede, ``Automatic bird sound detection in long real-field recordings: {Applications} and tools,'' \emph{Applied Acoustics}, vol.~80, pp. 1--9, 2014, publisher: Elsevier.

\bibitem{znidersic_using_2020}
E.~Znidersic, M.~Towsey, W.~K. Roy, S.~E. Darling, A.~Truskinger, P.~Roe, and D.~M. Watson, ``Using visualization and machine learning methods to monitor low detectability species—{The} least bittern as a case study,'' \emph{Ecological Informatics}, vol.~55, p. 101014, 2020, publisher: Elsevier.

\bibitem{sudharsan_ml-mcu_2021}
B.~Sudharsan, J.~G. Breslin, and M.~I. Ali, ``Ml-mcu: {A} framework to train ml classifiers on mcu-based iot edge devices,'' \emph{IEEE Internet of Things Journal}, vol.~9, no.~16, pp. 15\,007--15\,017, 2021, publisher: IEEE.

\bibitem{planque_xeno-canto_2005}
B.~Planqué, W.-P. Vellinga, S.~Pieterse, and J.~Jongsma, ``Xeno-canto: sharing bird sounds from around the world,'' \emph{Available: www. xeno-canto. org}, 2005.

\bibitem{piczak_esc_2015}
K.~J. Piczak, ``{ESC}: {Dataset} for environmental sound classification,'' in \emph{Proceedings of the 23rd {ACM} international conference on {Multimedia}}, 2015, pp. 1015--1018.

\bibitem{kahl_identifying_2019}
M.~S.~S. Kahl, ``Identifying birds by sound: large-scale acoustic event recognition for avian activity monitoring,'' Ph.D. dissertation, Chemnitz University of Technology, 2019.

\bibitem{park_specaugment_2019}
D.~S. Park, W.~Chan, Y.~Zhang, C.-C. Chiu, B.~Zoph, E.~D. Cubuk, and Q.~V. Le, ``Specaugment: {A} simple data augmentation method for automatic speech recognition,'' \emph{arXiv preprint arXiv:1904.08779}, 2019.

\bibitem{lin_mcunet_2020}
J.~Lin, W.-M. Chen, Y.~Lin, C.~Gan, S.~Han, and {others}, ``Mcunet: {Tiny} deep learning on iot devices,'' \emph{Advances in neural information processing systems}, vol.~33, pp. 11\,711--11\,722, 2020.

\bibitem{sandler_mobilenetv2_2018}
M.~Sandler, A.~Howard, M.~Zhu, A.~Zhmoginov, and L.-C. Chen, ``Mobilenetv2: {Inverted} residuals and linear bottlenecks,'' in \emph{Proceedings of the {IEEE} conference on computer vision and pattern recognition}, 2018, pp. 4510--4520.

\bibitem{deng_imagenet_2009}
J.~Deng, W.~Dong, R.~Socher, L.-J. Li, K.~Li, and L.~Fei-Fei, ``Imagenet: {A} large-scale hierarchical image database,'' in \emph{2009 {IEEE} conference on computer vision and pattern recognition}.\hskip 1em plus 0.5em minus 0.4em\relax Ieee, 2009, pp. 248--255.

\bibitem{deutel_combining_2024}
M.~Deutel, G.~Kontes, C.~Mutschler, and J.~Teich, ``Combining multi-objective bayesian optimization with reinforcement learning for tinyml,'' \emph{ACM Transactions on Evolutionary Learning}, vol.~5, no.~3, pp. 1--21, 2025.

\bibitem{deutel_energy-efficient_2023}
M.~Deutel, P.~Woller, C.~Mutschler, and J.~Teich, ``Energy-efficient {Deployment} of {Deep} {Learning} {Applications} on {Cortex}-{M} based {Microcontrollers} using {Deep} {Compression},'' in \emph{{MBMV} 2023; 26th {Workshop}}.\hskip 1em plus 0.5em minus 0.4em\relax VDE, 2023, pp. 1--12.

\bibitem{onnxruntime}
O.~R. developers, ``Onnx runtime,'' \url{https://onnxruntime.ai/}, 2021.

\bibitem{liu_visualizing_2019}
G.~Liu, H.~Zeng, and D.~K. Gifford, ``Visualizing complex feature interactions and feature sharing in genomic deep neural networks,'' \emph{BMC bioinformatics}, vol.~20, pp. 1--14, 2019, publisher: Springer.

\bibitem{zhu_learning_2022}
F.~Zhu, X.-Y. Zhang, R.-Q. Wang, and C.-L. Liu, ``Learning by seeing more classes,'' \emph{IEEE Transactions on Pattern Analysis and Machine Intelligence}, vol.~45, no.~6, pp. 7477--7493, 2022, publisher: IEEE.

\bibitem{dwd_sun}
{Deutscher Wetterdienst}, ``Global radiation in germany,'' \url{https://www.dwd.de/EN/ourservices/solarenergy/maps_globalradiation_mvs.html}.

\end{thebibliography}

\end{document}